\documentclass[11pt]{amii}
\usepackage{environ}
\newcommand{\figurepanel}[1]{(#1)}
\usepackage[T1]{fontenc}
\usepackage[utf8]{inputenc}
\usepackage{amsmath}
\makeatletter\@ifpackageloaded{newpxmath}{}{\usepackage{amssymb}}\makeatother
\usepackage{graphicx,booktabs,tabularx,array}
\usepackage{algorithm}
\usepackage{algpseudocode}
\usepackage{colortbl}
\usepackage{pifont}
\definecolor{rowhl}{gray}{0.92}      %
\definecolor{grouprow}{gray}{0.97}   %
\usepackage{xspace,xcolor}
\usepackage{enumitem}
\usepackage{placeins}
\usepackage{hyperref}
\usepackage{url}
\usepackage{cleveref}
\usepackage{tcolorbox}
\tcbuselibrary{skins,breakable,xparse}
\definecolor{promptbg}{HTML}{F6F8FB}
\definecolor{promptline}{HTML}{3B6EA5}
\tcbset{promptbreak/.style={breakable}}
\newtcolorbox[auto counter, number within=section,
              crefname={Prompt}{Prompts}, Crefname={Prompt}{Prompts}]{promptblock}[2][]{%
  enhanced, promptbreak,
  colback=promptbg, colframe=promptline, colbacktitle=promptline,
  fonttitle=\bfseries\small, fontupper=\small\ttfamily,
  boxrule=0.6pt, arc=1.5mm,
  left=2mm, right=2mm, top=1.2mm, bottom=1.2mm,
  before skip=6pt, after skip=6pt,
  title={Prompt \thetcbcounter: #2}, #1}
\usepackage{etoc}
\newcommand{\protege}{SWE-Prot\'eg\'e\xspace}
\newcommand{\ours}{UVA\xspace}            %
\newif\ifcolumnteaser
\newcommand{\narrowtablewidth}{\linewidth}
\providecommand{\figurepanel}[1]{{\fontsize{\dimexpr\linewidth*10/396\relax}{\dimexpr\linewidth*12/396\relax}\selectfont(#1)}}
\newcolumntype{Q}{>{\raggedright\arraybackslash}X}
\newcolumntype{P}[1]{>{\raggedright\arraybackslash}p{#1}}
\providecommand{\appendixcontents}{%
  \etocsettocdepth.toc{subsection}%
  \section*{Appendix Contents}%
  \begingroup
  \renewcommand{\contentsname}{}%
  \vspace{-2.2em}%
  \tableofcontents
  \endgroup
  \clearpage}


\usepackage{fontawesome5}
\urldef{\releaselink}\url{https://github.com/jwj1342/UVA}
\newcommand{\releasenote}{}
\renewcommand{\narrowtablewidth}{0.6\linewidth}
\RenewEnviron{abstract}{\global\let\abstractlist\BODY}
\title{Teaching Code Agents to Ask Without Leaking Private Code}
\author[1,2,3]{Wenjia~Jiang}
\author[1]{Jieke~Shi}
\author[1]{Zhensu~Sun}
\author[1]{Chengran~Yang}
\author[4,5]{Shidong~Pan}
\author[1]{Peixin~Zhang}
\author[6,7]{Yifan~Jia}
\author[8]{James~Hoang}
\author[8]{Xiwei~Xu}
\author[8]{Zhenchang~Xing}
\author[1]{David~Lo}
\author[2,3,\ddag,\dag]{Zhou~Yang}
\affiliation[1]{Singapore Management University}
\affiliation[2]{University of Alberta}
\affiliation[3]{Alberta Machine Intelligence Institute (Amii)}
\affiliation[4]{Columbia University}
\affiliation[5]{New York University}
\affiliation[6]{Singapore University of Technology and Design}
\affiliation[7]{AIDX TECH}
\affiliation[8]{Commonwealth Scientific and Industrial Research Organisation (CSIRO)}
\affiliation[\ddag]{Canada CIFAR AI Chair}
\affiliation[\dag]{Corresponding author}
\renewcommand\metadataformat[2][]{{\small \ifstrempty{#1}{}{{\bfseries #1:} }#2}}
\metadata{\faEnvelope~\email{zy25@ualberta.ca}, \email{wjjiang@smu.edu.sg}}
\metadata{\faGithub~\releaselink}
\date{\today}
\begin{document}
\begin{abstract}
Local coding agents can extend their reach by consulting a stronger cloud model. Such consultations,
however, often expose the repository's private identifiers, file paths, and source
code. Host-side masking hides recognized identifiers but leaves the surrounding code and the
agent's phrasing unchanged, while training for less disclosure risks suppressing consultation
or producing questions too vague to answer. We propose \emph{utility-verified abstraction} (UVA), which
teaches the agent to describe private entities by their role and type while retaining the technical
details an expert needs. From solved repair trajectories, we rewrite each question into this
form and keep the rewrite only if it discloses fewer private entities and the repair, replayed from the
same execution state with the rewrite, still succeeds. We fine-tune the agent on the verified
questions and ordinary repair trajectories. Across two
benchmarks, the trained agent consults more often than before training. On repository-level issue
repair, it solves $37.6\%$ of issues, against $25.3\%$ for the same model working alone, a gain from the
training as a whole rather than from consultation alone. Its questions
disclose $3.7$-fold fewer private entities and $1.4$-fold less verbatim code than the untrained agent's
questions to the same expert.\releasenote
\end{abstract}

\maketitle
\newcommand{\motivationcaption}{\textbf{One consultation, asked two ways.} A local coding agent works through a repair and,
at one step, consults a third-party cloud expert. \figurepanel{a} The untrained agent quotes three private
identifiers of the repository in its question. \figurepanel{b} Our trained agent describes the same
entities by role and type. The expert gives both the same advice and both repairs succeed, so the
abstract question gets the same help while the identifiers stay on the device.}
\newcommand{\motivationcolumncaption}{\textbf{One consultation, asked two ways.} A local coding agent works through a repair and,
at one step, consults a third-party cloud expert. In the top panel, the untrained agent quotes three
private identifiers of the repository in its question. In the bottom panel, our trained agent describes
the same entities by role and type. The expert gives both the same advice and both repairs succeed, so the
abstract question gets the same help while the identifiers stay on the device.}
\ifcolumnteaser
\begin{figure}[!t]
\centering
\includegraphics[width=\linewidth]{figures/motivation_column.pdf}
\caption{\motivationcolumncaption}
\vspace{-0.7cm}
\label{fig:motivation}
\end{figure}
\else
\begin{figure}[t]
\centering
\includegraphics[width=\linewidth]{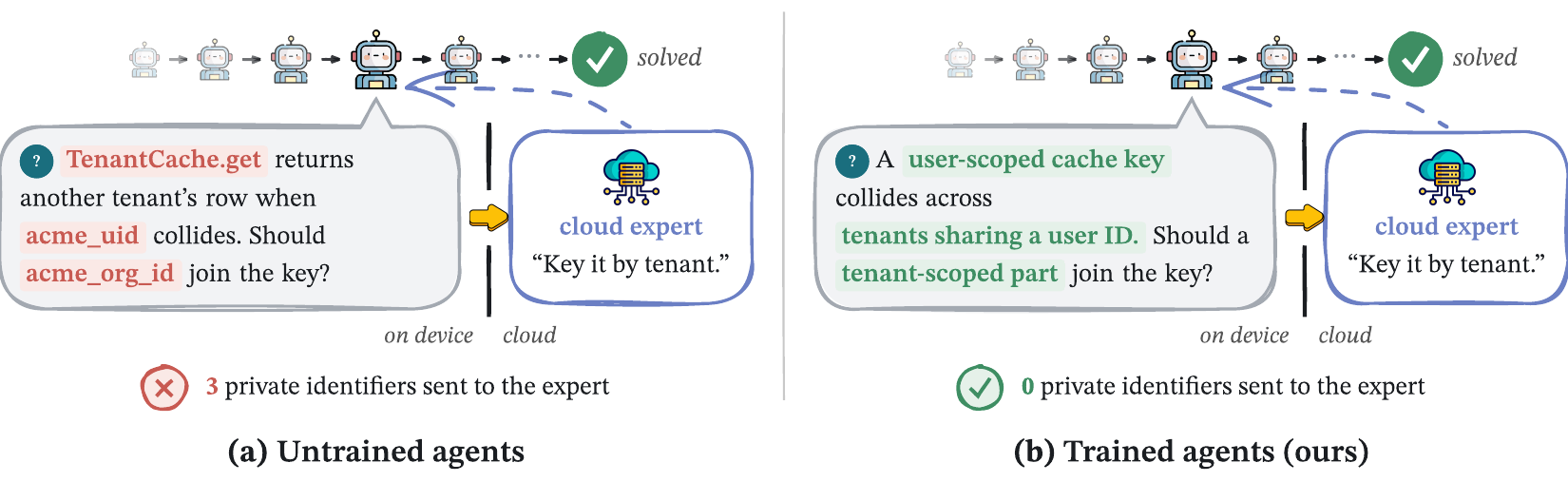}
\caption{\motivationcaption}
\label{fig:motivation}
\end{figure}
\fi

\etocsettocdepth.toc{none}
\section{Introduction}

Large language models (LLMs) have demonstrated strong capabilities in code generation and program repair~\citep{jiang2026survey,zhang2024autocoderover}. Coding agents, which integrate LLMs with development tools, can iteratively inspect files, modify code, and execute tests to produce a patch given an issue description and a repository~\citep{swebench2024,sweagent2024}. Their growing capabilities have led to increasing adoption in real-world software development~\citep{10.1145/3822180}. However, many  coding agents rely on proprietary LLMs hosted by external providers, which raises privacy concerns when repository content is sent outside the organization. Running a smaller model locally can keep repository operations within the organization, but may limit the agent's ability to resolve complex issues. To address this trade-off, local execution can be combined with remote consultation. A small \emph{student} model runs on the machine that holds the repository and performs all editing and testing, while a larger \emph{expert} model provides advice remotely without directly accessing the repository. Prior work has explored how to allocate computation across models through cascades and routing~\citep{frugalgpt2024,routellm2025}, and \protege shows that a local student can learn when to consult a stronger expert~\citep{sweprotege2026}.

However, keeping repository operations local does not fully address these privacy concerns. The student may still include private identifiers, file paths, or source code in its consultations, making this information visible to the external provider. Existing approaches can limit such disclosure through prompt-based redaction and abstraction~\citep{papillon2025,dou2024selfdisclosure}, which instruct the model to remove or generalize private information before the question is sent to an external model. The student must nevertheless determine which details can be removed without obscuring the problem, and a privacy instruction alone provides no task-level feedback on whether the resulting question retains the information needed for useful advice. Such an instruction may also lower the consultation rate rather than improve the questions, in which case a reduction in total disclosure reflects silence rather than safer consultation. Alternatively, a host-side sanitizer can replace recognized private strings with placeholders before transmitting the question. This hides the identifiers that the sanitizer recognizes, but the surrounding source code is still transmitted. String replacement alone also does not check whether the sanitized question preserves the context the expert needs to give useful repair advice. \textbf{The central challenge is therefore to reduce the private information disclosed in a consultation while preserving the technical content needed for useful advice.} Removing too little leaves repository content exposed, while removing too much may omit the conditions that determine the correct repair. Training the student to make these choices requires examples of questions that satisfy both requirements. Whether a rewritten question remains useful, however, can be judged only after the expert has answered it and the student has completed the repair, so such examples cannot be labeled from the question text alone.

In this paper, we present \emph{utility-verified abstraction} (\ours), a training procedure that teaches local coding agents to formulate useful questions with less private information. \textbf{\ours constructs training examples from the consultations of successful repair trajectories, retaining an abstract rewrite of a question only if it discloses fewer private entities and the replayed repair still succeeds.} An offline teacher rewrites each original question given its interaction history, describing private entities by their roles and types while preserving the symptom, the relevant technical constraints, and the decision on which advice is sought (Figure~\ref{fig:motivation}). Each rewrite then undergoes two checks. First, we count the private entities in the original and rewritten questions, namely the identifiers, file and module names, and paths that the student has read locally and that are not standard-library or public names. Second, we restore the repository snapshot and interaction history at the original consultation, obtain the expert's answer to the rewritten question, and resume the student's repair with that answer under the same remaining budget. A rewrite enters training only if it discloses strictly fewer private entities and the replayed repair passes the task's tests. We then use LoRA-based supervised fine-tuning to train the student to generate each accepted rewrite from its original interaction history, mixing these examples with full repair trajectories to help preserve its repair capability. At deployment, the student decides when to consult and generates its own questions, without the offline teacher or any host-side rewriting.

We conduct experiments on SWE-bench Verified and Terminal-Bench using Qwen3.5-9B \citep{qwen3.5} as the student. With DeepSeek-V3.2 \citep{deepseekai2025deepseekv32} as the expert for both \ours and the untrained consulting baseline, \ours improves $\mathrm{pass}@1$ on SWE-bench Verified from $24.6\%$ to $37.6\%$, while reducing mean disclosure per consultation from $4.42$ to $1.20$ private entities and from $8.4$ to $5.9$ verbatim source-code $5$-grams. These reductions are not obtained by consulting less, as the consultation rate rises from $4.8\%$ to $7.4\%$. The gain in solve rate is attributable to the training as a whole rather than to consultation alone, as the trained student reaches $36.2\%$ without an expert. Compared with the host sanitizer, \ours achieves a higher solve rate ($37.6\%$ versus $25.0\%$), although its private-entity disclosure remains higher ($1.20$ versus $0.97$). With Claude Opus 4.8 \citep{anthropic2026opus48} as the expert, the trained student achieves $40.6\%$ $\mathrm{pass}@1$ on SWE-bench Verified and $43.8\%$ accuracy on Terminal-Bench. In ablations with Claude Opus 4.8 as the expert, removing the repair replay lowers $\mathrm{pass}@1$ from $40.6\%$ to $22.4\%$, below the same model working alone. We also observe \emph{privacy by silence} on SWE-bench Verified, where a privacy prompt reduces the consultation rate without reducing mean disclosure per consultation. This finding indicates that evaluations of private consultation should report the consultation rate together with the disclosure per consultation. Our privacy claims concern literal disclosure and do not establish protection against information inferred from the transmitted content.

Our contributions are summarized as follows:
\begin{itemize}[leftmargin=1.4em,itemsep=2pt,topsep=3pt,parsep=0pt]
\item We propose \ours, a training procedure that builds supervision for consulting a cloud expert with less private information, keeping only role-and-type rewrites that pass a disclosure check and a repair replay.
\item We identify \emph{privacy by silence} on SWE-bench Verified, showing that a lower consultation rate can accompany an apparent privacy improvement without lower mean disclosure per consultation.
\item We evaluate \ours on two benchmarks against no consultation, unrestricted consultation, a privacy prompt, and a deterministic host sanitizer, and release the source code.
\end{itemize}
\section{Related Work}
\label{sec:related}

Small and large models are usually combined by choosing which model answers, either by cascading on a
confidence signal \citep{frugalgpt2024,automix2024} or by routing before generation
\citep{routellm2025,hybridllm2024}. A second line checks a message before it is sent, with guard models
that classify the message \citep{inan2023llamaguard,rebedea2023nemo} or sanitizers that substitute its
sensitive spans \citep{chen2023has,pilan2022tab}. Both lines leave unchanged the policy that writes the
message. \protege trains an on-device repair agent, which alone acts on the repository, to decide when
to escalate to an expert \citep{sweprotege2026}, but it does not control what the question discloses.
PAPILLON optimises the prompt sent to the remote model \citep{papillon2025},
\citet{huang2026needtoknow} train a rewriter on spans annotated as task-essential,
\citet{dou2024selfdisclosure} rephrase private content into a less specific form, and
\citet{zeng2026seekhelp} concurrently train an on-device model to decide when to escalate and how to
phrase the request. Scrouting verifies its outbound handoff by checking whether the claims in it are
true \citep{bhola2026scrouting}. None of these methods tests a rewritten request by re-running the
task. On the training side, tool calls are learned by imitating a teacher's traces
\citep{qin2024toolllm,patil2024gorilla} or by rewarding executed calls \citep{qiao2024trice,jin2025searchr1}.
Rejection-sampling fine-tuning instead keeps a sampled candidate only when it passes an acceptance
criterion, a likelihood margin in Toolformer \citep{schick2023toolformer} and task success in TALM and
STaR \citep{parisi2022talm,zelikman2022star}; this filter has since been applied to unit-tested code and
repository trajectories \citep{haluptzok2023selftaught,pan2025swegym} and, closest to our setting, to
query rewriters \citep{ma2023query} and to models that write the question they send
\citep{andukuri2024stargate,ambati2025socratic}. These criteria measure utility alone and discard the
rejected candidates or use them as negatives for preference learning \citep{dpo2023}, whereas ours
keeps a rewrite only if the replayed repair succeeds and the rewrite discloses strictly fewer private
entities than the original question (Appendix~\ref{app:related}).

\begin{figure*}[!t]
\centering
\includegraphics[width=\linewidth]{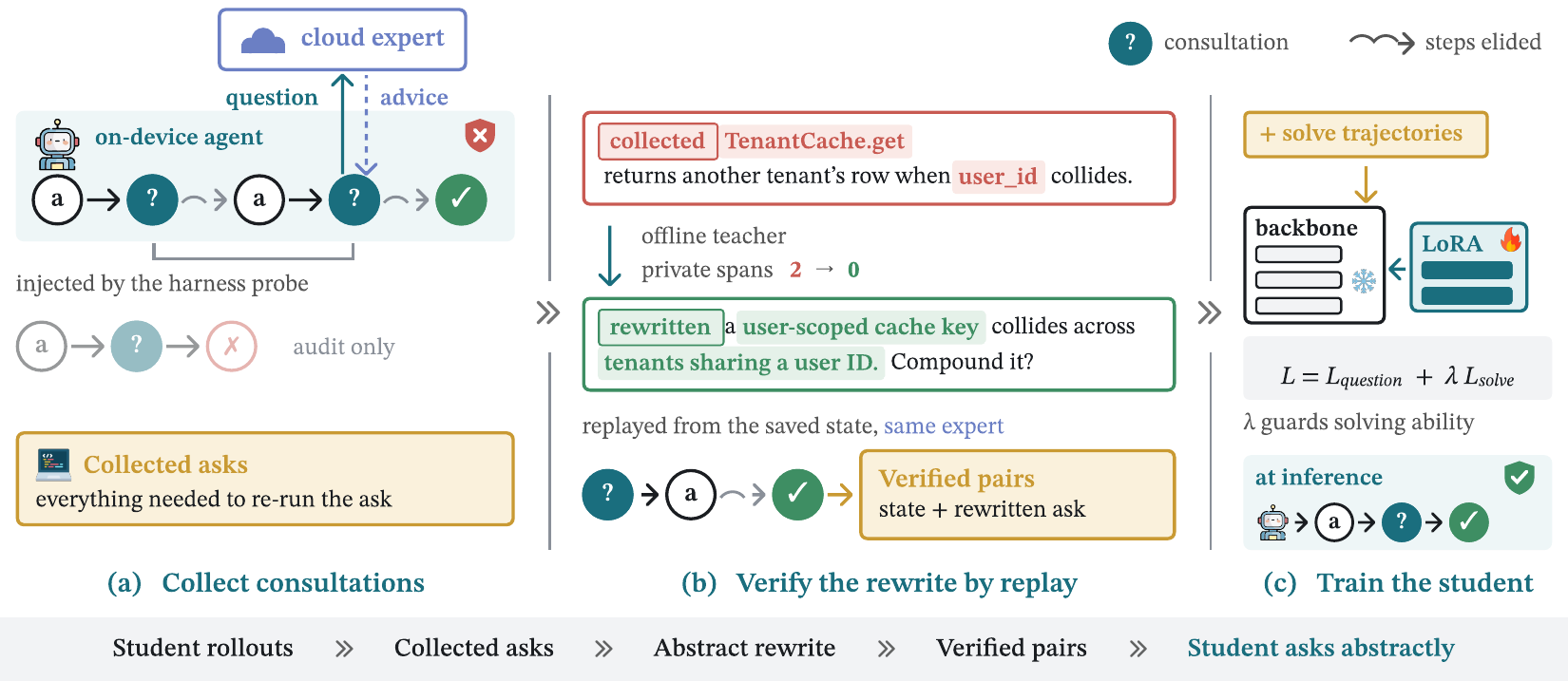}
\caption{\textbf{Overview of \ours.}
\figurepanel{a} Consultations are collected from successful student rollouts.
\figurepanel{b} An offline teacher describes private entities by their roles and types. A rewrite is retained only if it strictly reduces measured disclosure and the repair passes the task's tests when replayed from the saved state.
\figurepanel{c} The student is fine-tuned with LoRA on the verified pairs together with repair trajectories. At inference, it generates consultations directly without a rewriter.
The example and disclosure counts are illustrative.}
\vspace{-0.7cm}
\label{fig:pipeline}
\end{figure*}
\section{Problem Setting}
\label{sec:setting}

We consider a repair task consisting of an issue description, a repository, and an evaluation test suite. A small student model runs locally with policy $\pi_\theta$, selecting each action from its interaction history $h$. The student may also consult a remote expert $E$, which receives only a question $q$ and a fixed system prompt, while all repository operations remain local. A complete run is a rollout $\tau$, with $S(\tau)=1$ if its submitted patch passes the required tests and $S(\tau)=0$ otherwise.

\paragraph{Threat model.}
The local host, the student, and the offline teacher that rewrites training questions are trusted. The
teacher reads the original questions, so on a private repository it must run inside the trust boundary.
The expert's provider is honest but curious and sees every transmitted question. Inference from the
transmitted content lies outside this scope.

\paragraph{Disclosure measures.}
We measure the literal disclosure of private repository content, which we call \textbf{P1} to separate
it from hard secrets such as credentials (\textbf{P0}). Both classes are determined automatically (Appendix~\ref{app:metrics}). Let the reference $r_h$
be the file contents, tracebacks, and command outputs that the student has observed up to $h$. An
entity is an identifier, a module or file name, or a path. Let
$I(r_h)$ be the set of private entities, those that appear in $r_h$ and are not on a versioned allowlist
of standard-library and public names. Let $W_5(r_h)$ be the set of token $5$-grams on the lines of $r_h$
that look like source code, with comments removed. For a transmitted question $q$, we define the leaked
private-entity count $L_{\mathsf{P1}}$
and the outbound code overlap $\mathsf{OCO}$ as
\begin{align}
L_{\mathsf{P1}}(q;r_h)
&=
\bigl|\{u \in I(r_h):{}\notag\\
&\qquad u \text{ occurs verbatim in } q\}\bigr|,\label{eq:lp1}\\
\mathsf{OCO}(q;r_h)
&=
\left|W_5(r_h) \cap w_5(q)\right|,\label{eq:oco}
\end{align}
where $w_5(q)$ is the set of token $5$-grams over all lines of $q$. Both are computed on the actual transmitted text, including any host-side transformation. Appendix~\ref{app:metrics} specifies the reference construction and filtering rules. We omit $r_h$ when clear from context.

Our objective is to improve repair success while reducing disclosure per consultation. For a fixed expert and execution budget, we report the solve rate together with the mean disclosure per transmitted question, the total disclosure over all transmitted questions divided by their number. This mean is undefined for a policy that never consults, so Table~\ref{tab:main} marks it n/a for \emph{Solo}. Because a policy can also lower its total disclosure per task by consulting less, we report the consultation rate, the share of rollouts that consult at least once, to separate lower disclosure per question from less frequent consultation. Appendix~\ref{app:metrics} states these quantities formally.
\section{Method}
\label{sec:method}

\textbf{Utility-Verified Abstraction} (\ours) trains the student to formulate consultations that disclose less private information while retaining the technical content needed for repair. As shown in Figure~\ref{fig:pipeline}, the procedure collects consultation candidates, constructs abstract rewrites verified by disclosure checks and repair replay, and fine-tunes the student on the accepted examples.

\subsection{Collecting unconstrained consultations}

We collect consultations from student rollouts on training tasks with executable tests, without imposing privacy constraints on the generated questions. The student consults the expert by issuing a command, in the same format as its local tool calls. To collect consultations from different stages of a repair, we prompt the student to consult at scheduled steps, starting at a fixed step and repeating at a fixed interval up to a per-rollout cap (Appendix~\ref{app:config}). At each scheduled step, we append an instruction to consult to a temporary copy of the interaction history and sample a question from the student on this copy. We send the question to the expert and add the expert's response to the running trajectory, which never contains the appended instruction. We use this schedule only for data collection; during evaluation, the student decides when to consult.

We keep consultations only from successful rollouts, those with $S(\tau)=1$, so that each one comes from a repair that succeeded. This filter does not establish that the consultation caused the success. We call the kept consultations candidates and denote their set by $\mathcal C$. We keep one candidate per identical interaction history, cap the number of candidates per task, and keep each candidate and its replay within the data split of its originating task. Each candidate records the interaction history $h$, the original question $q^{-}$, the expert response, the repository snapshot, the random seed, and the remaining execution budgets, which together allow replay from the saved consultation state.

\subsection{Verifying rewrites by repair replay}

For each candidate $(h,q^{-})\in\mathcal C$, an offline teacher $R$ generates an abstract rewrite $q^{+}=R(h,q^{-})$. The teacher is distinct from the expert $E$ and is used only during data construction. Its instruction is to replace private entities with descriptions of their roles and types while preserving the observed symptom, relevant technical constraints, and decision requiring advice. Simply deleting details can remove conditions that determine the correct repair, while replacing names with opaque placeholders such as \texttt{IDENT\_3} does not tell the expert what role each entity plays. For example, an abstract question can describe a user-scoped cache key that collides across tenants and ask whether the key should include a tenant-scoped component. This retains the technical condition and proposed decision without exposing repository-specific names. Appendix~\ref{app:teacher} provides the teacher prompt.

Before replay, a rewrite must pass a fixed privacy predicate $A$, disclose strictly fewer private entities than the original question, and pass a validity check $V$. The privacy predicate is
\begin{equation}
A(q)
\iff
L_{\mathsf{P0}}(q)=0
\;\wedge\;
\hat L_{\mathsf{P1}}(q)\leq\kappa_{\mathsf{P1}},
\label{eq:accept}
\end{equation}
where $\wedge$ denotes logical \emph{and} and $L_{\mathsf{P0}}(q)$ counts hard secrets found by a pattern scan, such as credentials, internal URLs, and private absolute paths. The detector-only count $\hat L_{\mathsf{P1}}(q)$ counts every entity span that the entity recognizer flags in $q$, whether or not the span appears in $r_h$, and $\kappa_{\mathsf{P1}}=2$. The predicate $A$ therefore depends on the question alone, while the strict reduction compares both questions against $r_h$. Accepted training rewrites contain no pattern-detected hard secrets and at most two flagged P1 spans. The validity check $V(q^{+})$ requires a nonempty question of bounded length in plain prose, with no code fence, shell command, or opaque placeholder, whose closing sentence poses the question. The allowed abstraction structure is specified in Appendix~\ref{app:teacher}.

For candidates that pass these checks, we restore the repository snapshot, interaction history, and remaining budget at the original consultation. The expert answers $q^{+}$, and the student resumes under the original decoding and retry rules. Let $\tau^{+}$ denote the resulting rollout. During collection and replay, we send each question to the expert unmodified, so the scored question is exactly the text that the student or teacher wrote. A candidate is accepted only when
\begin{equation}
\begin{aligned}
&G(h,q^{+},q^{-}) \iff A(q^{+})\\
&\quad{}\wedge L_{\mathsf{P1}}(q^{+})<L_{\mathsf{P1}}(q^{-}) \wedge V(q^{+})\\
&\quad{}\wedge S(\tau^{+})=1.
\end{aligned}
\label{eq:gate}
\end{equation}
Both questions are scored against the same reference $r_h$. The strict reduction applies to $L_{\mathsf{P1}}$, and $\mathsf{OCO}$ is recorded for analysis only. We call the last condition, $S(\tau^{+})=1$, the replay gate.

The procedure thus keeps a rewrite only if it discloses fewer private entities than the original question and the repair that continues from the expert's answer to it succeeds. Replaying from the saved state keeps the history before the consultation identical, but a passing replay does not establish that the consultation was necessary or that the rewrite preserves the probability of success. Rejected candidates are recorded by rejection stage (Appendix~\ref{app:pairs}).

\subsection{Training strategy}

The accepted pairs form the supervision set
$\mathcal D=\{(h,q^{+}):(h,q^{-})\in\mathcal C,\ G(h,q^{+},q^{-})\}$.
We use LoRA-based supervised fine-tuning to train the student to generate each accepted question $q^{+}$ from its original interaction history $h$, keeping the backbone parameters frozen. Because the same adapter is active throughout the agent loop, question supervision can also affect ordinary repair actions. We therefore combine the consultation examples with full repair trajectories, minimizing
\begin{equation}
\mathcal L(\theta)
=
\mathcal L_{\mathsf{question}}(\theta;\mathcal D)
+
\lambda\,
\mathcal L_{\mathsf{solve}}(\theta;\mathcal D_{\mathsf{solve}}).
\label{eq:loss}
\end{equation}
Here, $\mathcal L_{\mathsf{question}}$ is the negative log-likelihood of the rewritten question conditioned on $h$, and $\mathcal L_{\mathsf{solve}}$ is the negative log-likelihood of the student's actions in the repair trajectories $\mathcal D_{\mathsf{solve}}$. The trajectories in $\mathcal D_{\mathsf{solve}}$ are action steps sampled from the same successful collection rollouts, with the consultation turns excluded from the targets, so an original question $q^{-}$ appears only as context. The repair trajectories preserve the student's repair capability while the student learns to write abstract questions. Because their targets are the student's own successful repair actions, some taken after an expert reply had entered the context, training on them can also raise the solve rate of rollouts that never consult (Appendix~\ref{app:decomp}).
\section{Experiments}
\label{sec:experiments}

We evaluate \ours on two benchmarks in terms of task success and outbound disclosure, and examine the effects of its training components and the amount of supervision.

\subsection{Setup}

\label{sec:setup}

\paragraph{Benchmarks.}
Our primary benchmark is \textbf{SWE-bench Verified} \citep{swebench2024}, which contains 500 human-validated GitHub issues with per-instance execution environments. We score patches with a local scorer that follows the official rule, under which a patch resolves an issue when all fail-to-pass tests pass and no pass-to-pass test fails. To evaluate beyond program repair, we additionally use all 89 tasks in \textbf{Terminal-Bench 2.0} \citep{terminalbench2025}, with success determined by its task-specific verifiers.

\paragraph{Baselines.}
We compare \ours with four baselines, all of which use the same untrained student.
\emph{Solo} runs the untrained student without consultation and shows how many tasks the student solves on its own.
\emph{Natural} lets the untrained student consult the expert and sends its questions unmodified.
\emph{Privacy prompt} adds the offline teacher's abstraction instruction verbatim to the untrained student's system prompt (Appendix~\ref{app:teacher}), without updating model parameters. The instruction asks the student to describe private entities by their roles and types while preserving the technical conditions and repair decision.
\emph{Host sanitizer} applies a deterministic transform $T$ that replaces the identifiers, module names, paths, URLs, and secrets it recognizes with placeholders. The expert receives $T(q)$, and we map the placeholders in its response back to the original strings locally.
\emph{Ours} uses the student trained by \ours. Natural, Privacy prompt, and Ours transmit their generated questions without host-side sanitization.

\paragraph{Metrics.}
We report \textbf{\boldmath$\mathrm{pass}@k$} on SWE-bench Verified using the standard unbiased estimator from $n\geq k$ samples per instance \citep{chen2021codex}, so that $\mathrm{pass}@1$ estimates the solve probability, and $\mathrm{pass}@1$ on Terminal-Bench, its native single-attempt accuracy.
\textbf{Ask\%} is the percentage of rollouts with at least one transmitted consultation.
\textbf{\boldmath$L_{\mathsf{P1}}$} counts the distinct private entities in a transmitted question and \textbf{\boldmath$\mathsf{OCO}$} counts the distinct source-code $5$-grams that the question copies verbatim, both averaged over all transmitted questions (Appendix~\ref{app:metrics}). Metric definitions, statistical procedures, and consultation accounting are provided in Appendices~\ref{app:metrics}, \ref{app:stats}, and~\ref{app:results}.

\begin{table*}[t]
\centering
\footnotesize
\setlength{\tabcolsep}{6pt}
\renewcommand{\arraystretch}{1.18}
\begin{tabular}{l ccccc !{\color{black!35}\vrule} cccc}
\toprule
& \multicolumn{5}{c}{\textbf{SWE-bench Verified}} & \multicolumn{4}{c}{\textbf{Terminal-Bench}} \\
\cmidrule(lr){2-6}\cmidrule(l){7-10}
Policy & $\mathrm{pass}@1\uparrow$ & $\mathrm{pass}@3\uparrow$ & Ask\% & $L_{\mathsf{P1}}\downarrow$ & $\mathsf{OCO}\downarrow$ & $\mathrm{pass}@1\uparrow$ & Ask\% & $L_{\mathsf{P1}}\downarrow$ & $\mathsf{OCO}\downarrow$ \\
\midrule
Solo (no ask) & 25.3 & 31.4 & \;0 & n/a & n/a & 20.2 & \;0 & n/a & n/a \\
Natural & 24.6 & 32.4 & 4.8 & 4.42 & 8.4 & 24.7 & 7.9 & 4.18 & 4.0 \\
Privacy prompt & 23.8 & 31.2 & 1.4 & 4.64 & 9.5 & 20.2 & 7.9 & 3.17 & 2.3 \\
Host sanitizer & 25.0 & 28.6 & 4.6 & 0.97 & 6.0 & 22.5 & 5.6 & 0.71 & 3.6 \\
\addlinespace[2pt]
\rowcolor{rowhl}\multicolumn{10}{l}{\emph{\textbf{Ours}, by cloud expert}} \\
\rowcolor{rowhl}\quad DeepSeek-V3.2 & 37.6 & 45.6 & 7.4 & 1.20 & 5.9 & 29.2 & 9.0 & 2.35 & 1.4 \\
\rowcolor{rowhl}\quad GPT-5.6 Sol & 38.3 & 48.8 & 8.2 & 2.14 & 6.2 & 36.0 & 7.9 & 1.97 & 2.0 \\
\rowcolor{rowhl}\quad Claude Opus 4.8 & 40.6 & 59.2 & 7.6 & 2.31 & 4.8 & 43.8 & 11.2 & 1.36 & 2.1 \\
\bottomrule
\end{tabular}
\caption{\textbf{Solve rate and outbound disclosure of each consultation policy.}
The first four rows are the baselines, all run with the untrained student, and the \emph{Ours} rows
are the student trained by \ours. The baselines solve about as many tasks with an expert as without
one. With the same expert, the trained student solves more tasks and consults more often. Each of its
questions also discloses fewer private entities and less verbatim code. Its private-entity count stays
above the host sanitizer's.}
\label{tab:main}
\vspace{-0.3cm}
\end{table*}

\paragraph{Implementation details.}
We build the consultation training pairs from synthetic repair tasks in \textbf{SWE-smith} \citep{swesmith2025}. The student is Qwen3.5-9B \citep{qwen3.5}, adapted using rank-$16$ LoRA on its linear projections. All consulting policies use DeepSeek-V3.2 \citep{deepseekai2025deepseekv32} as the expert unless otherwise specified; we additionally evaluate \ours with GPT-5.6 Sol \citep{openai2026gpt56} and Claude Opus 4.8 \citep{anthropic2026opus48}. Each task allows at most three consultations. Model versions, prompts, consultation limits, execution budgets, privacy rules, sample counts, and random seeds are fixed before the final evaluation (Appendix~\ref{app:config}). All benchmark instances remain in the evaluation denominator, with unscorable instances counted as failures. The agent's shell activates the per-instance test environment that the grader uses.

\subsection{Results}
\label{sec:passk}
\paragraph{Task solving.}
The untrained student solves about as many tasks with an expert as without one. On Verified, its
$\mathrm{pass}@1$ is $25.3$ under \emph{Solo}, $24.6$ under \emph{Natural}, $23.8$ under
\emph{Privacy prompt}, and $25.0$ under \emph{Host sanitizer}, within $1.5$ points of one another. On
Terminal-Bench, no consulting policy differs significantly from \emph{Solo} under the exact McNemar
test on paired tasks, with $p=0.34$ for \emph{Natural}, $1.00$ for \emph{Privacy prompt}, and $0.75$
for \emph{Host sanitizer}. With the same expert, the student trained by \ours solves more tasks. With
DeepSeek-V3.2 as the expert, its $\mathrm{pass}@1$ on Verified is $37.6$, against $24.6$ for
\emph{Natural} and $25.3$ for \emph{Solo}, and its $\mathrm{pass}@3$ is $45.6$, against $32.4$ for
\emph{Natural}. On Terminal-Bench it solves $26$ of the $89$ tasks, against $22$ for \emph{Natural}
and $18$ for \emph{Solo}. A $4$-task difference cannot reach significance under this test (its smallest possible $p$ is
$0.125$), so Terminal-Bench alone does not separate \ours from \emph{Natural} with this expert. With GPT-5.6 Sol as the expert, the trained
student reaches $38.3$ on Verified and $36.0$ on Terminal-Bench, and with Claude Opus 4.8 it reaches
$40.6$ and $43.8$. We did not run the untrained student with these experts, so their gain over \emph{Natural} mixes
the effect of the questions with that of a stronger expert.

\begin{figure*}[t]
\centering
\begin{minipage}[t]{0.33\linewidth}\centering
\includegraphics[width=\linewidth]{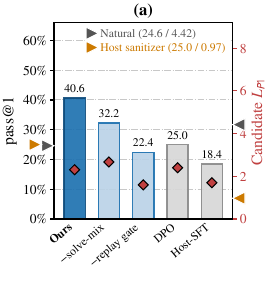}
\end{minipage}\hfill
\begin{minipage}[t]{0.33\linewidth}\centering
\includegraphics[width=\linewidth]{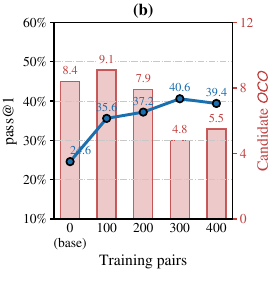}
\end{minipage}\hfill
\begin{minipage}[t]{0.33\linewidth}\centering
\includegraphics[width=\linewidth]{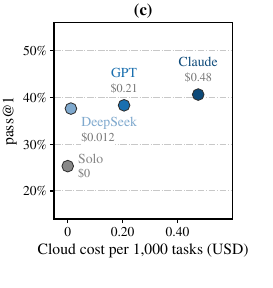}
\end{minipage}
\vspace{-0.7cm}
\caption{\textbf{Ablation on training strategy and training data, and the overall costs.} In
\figurepanel{a} and \figurepanel{b}, the trained students consult Claude Opus 4.8, while the marks taken
from Table~\ref{tab:main} and the zero-pair point are the untrained student with DeepSeek-V3.2.
\figurepanel{a} Disclosure is $L_{\mathsf{P1}}$; the triangles mark \emph{Natural} and \emph{Host
sanitizer}. \figurepanel{b} Nested budgets of $100$--$400$ accepted pairs; disclosure is
$\mathsf{OCO}$. \figurepanel{c} Cloud consultation cost per 1,000 tasks against the solve rate, each
point labelled with its price.}
\vspace{-0.3cm}
\label{fig:analysis}
\end{figure*}

\paragraph{Collaboration quality.}
A private consultation policy is effective only if it keeps consulting and sends less in each
consultation, and each baseline gives up one of the two. The privacy prompt lowers the total
disclosure per task by consulting less. On Verified it consults in $7$ of $500$ rollouts, against $72$
of $1{,}500$ for \emph{Natural}, a gap significant under Fisher's exact test at $p=3\times10^{-4}$,
while its $14$ questions disclose slightly more on average, $4.64$ private entities against $4.42$ and
$9.5$ source-code $5$-grams against $8.4$. We call this \emph{privacy by silence}, an apparent privacy
gain that comes from asking less rather than from disclosing less in each question. Its Verified Ask\% and disclosure, like
those of \emph{Host sanitizer}, come from one of three runs. On Terminal-Bench the privacy prompt consults as often as \emph{Natural}, in
$7.9\%$ of tasks, and lowers $L_{\mathsf{P1}}$ from $4.18$ to $3.17$, so privacy by silence appears
only on Verified. The host sanitizer keeps consulting and replaces the names it recognizes, but it
forwards the surrounding code, $6.0$ source-code $5$-grams per question on Verified against $8.4$ for
\emph{Natural}. \ours keeps consulting and sends less in each consultation. With DeepSeek-V3.2 as the
expert, the trained student consults more often than the untrained student, in $7.4\%$ of Verified
rollouts against $4.8\%$ and in $9.0\%$ of Terminal-Bench tasks against $7.9\%$. Each of its questions
discloses $3.7$-fold fewer private entities, $1.20$ against $4.42$, and $1.4$-fold less verbatim code,
$5.9$ against $8.4$, on Verified, and $1.8$-fold fewer entities and $2.9$-fold less code on
Terminal-Bench. Without any host transform, its verbatim-code overlap matches or undercuts the host
sanitizer's on both benchmarks, at $5.9$ against $6.0$ and $1.4$ against $3.6$
(Appendix~\ref{app:hostcompare}), while its private-entity count stays above the sanitizer's, at $1.20$
against $0.97$ on Verified. With the other experts, the trained student's disclosure per question also
stays below that of \emph{Natural} with DeepSeek-V3.2. Training changes how often the student consults
but not when, since the trained student consults at a median step of $39$ against $38$ for
\emph{Natural} (Appendix~\ref{app:timing}).

\paragraph{Ablation on the training strategy.}
\label{sec:mechanism}\label{sec:ablations}
We remove or replace one component of \ours at a time, retrain each variant on the same candidate
pool, initialization, and update budget as the $300$-pair student of Table~\ref{tab:main}, and evaluate
it on Verified, as in every panel of Figure~\ref{fig:analysis} (panel \figurepanel{a}). The four variants drop the repair-trajectory mixture, drop
the replay gate, replace supervised fine-tuning on $q^{+}$ with DPO on the contrast between $q^{+}$ and
$q^{-}$, and train toward the deterministic sanitizer's output instead of the abstraction, which we call
\emph{Host-SFT}. All trained students in panels \figurepanel{a} and \figurepanel{b} consult Claude Opus
4.8 \citep{anthropic2026opus48}, while the marks taken from Table~\ref{tab:main} come from the untrained
student with DeepSeek-V3.2 \citep{deepseekai2025deepseekv32} or with no expert, so a comparison with
those marks also changes the expert.

Every trained variant discloses between $1.60$ and $2.67$ private entities per question, below the
$4.42$ of \emph{Natural}, and the variants differ mainly in how many tasks they solve. Without the
repair-trajectory mixture, the student still discloses only $2.67$ entities, but $\mathrm{pass}@1$
falls from $40.6$ to $32.2$, so the question supervision alone lowers disclosure and the repair
trajectories are needed to keep the solve rate. Without the replay gate, disclosure falls furthest, to
$1.60$ entities, and $\mathrm{pass}@1$ falls to $22.4$, below the $25.3$ of \emph{Solo}. This is
consistent with the replay gate removing rewrites that omit details the repair depends on, which the
student would otherwise learn to omit. DPO reaches the disclosure of \ours, $2.40$ against $2.31$, but
solves only $25.0$, so training on the contrast with $q^{-}$ lowers disclosure without the solve gain of
supervised fine-tuning on $q^{+}$. Host-SFT solves $18.4$ and discloses $1.70$, more than the $0.97$ of
the sanitizer it imitates, so imitating placeholders reaches neither the sanitizer's disclosure nor the
solve rate of abstraction. The full procedure gives the highest solve rate of the five at a disclosure
within the range of the variants, so removing or replacing any component lowers the solve rate.

\paragraph{Ablation on the training data.}
To measure how much training data the abstraction needs, we retrain the student on the first $100$,
$200$, $300$, and $400$ accepted pairs, each subset containing the previous one, under the recipe of
Eq.~\ref{eq:loss} with the mixture ratio and epoch count held fixed
(Figure~\ref{fig:analysis}\figurepanel{b}). The point at zero pairs, marked base, is \emph{Natural}.
Because the mixture ratio is fixed, $|\mathcal D_{\mathsf{solve}}|$ grows with the pair budget, so the
curve varies both terms of Eq.~\ref{eq:loss} together rather than isolating the consultation pairs.
Table~\ref{tab:main}'s \emph{Ours} rows use the $300$-pair student (Appendix~\ref{app:config}). This
panel reads disclosure as $\mathsf{OCO}$ for two reasons. Renaming symbols leaves the surrounding code
in place, so $\mathsf{OCO}$ separates an abstract question from a pseudonymized one, and $\mathsf{OCO}$
needs no entity recognizer. Up to $300$ pairs, $\mathrm{pass}@1$ rises to $40.6$, while $\mathsf{OCO}$
stays near the $8.4$ of \emph{Natural} at $100$ pairs and then falls to $4.8$, below the host
sanitizer's $6.0$. This is consistent with the abstraction being learned from the pairs rather than
from the instruction alone, since the privacy prompt gives the untrained student the same instruction
and does not lower $\mathsf{OCO}$ ($9.5$ against $8.4$ in Table~\ref{tab:main}). At $400$ pairs,
$\mathrm{pass}@1$ is $39.4$ and $\mathsf{OCO}$ is $5.5$, so collecting more than $300$ pairs brings no
further gain under this recipe.

\paragraph{Cost analysis.}
Figure~\ref{fig:analysis}\figurepanel{c} plots, for each expert, the cost of cloud consultation per
$1{,}000$ tasks against the solve rate, with the student's own inference counted separately.
Consultation costs \$$0.012$ per $1{,}000$ tasks with DeepSeek-V3.2, \$$0.21$ with GPT-5.6 Sol, and \$$0.48$
with Claude Opus 4.8,
under $0.2\%$ of the about \$$268$ that the student's own inference costs. Because the student runs on
local hardware, we price its inference at OpenRouter's published per-token rate for Qwen3.5-9B.
Appendix~\ref{app:distributions} gives the measured length and the pricing of a consultation.
Training is a one-off cost
that amortizes as the deployed student is used, so we do not account for it here.

\section{Conclusion}
This paper shows that a local code agent can learn to consult a cloud expert while exposing less of
its repository. Trained only on rewrites that disclose less and still lead to a successful replayed
repair, the student solves $9$ to $24$ points more tasks than the same model working alone on both
benchmarks, consults at least as often as before training, and discloses fewer private entities and
less verbatim code per question. Removing any training component lowers the solve rate.

\FloatBarrier
\label{main-end}

\clearpage
\bibliographystyle{assets/plainnat}
\bibliography{ref}
\clearpage
\appendix
\appendixcontents
\section{Extended Related Work}
\label{app:related}
Organised by mechanism rather than topic, because the distinction our method turns on is mechanical.
Almost every established answer acts on the traffic from outside the policy, whereas ours acts on the policy
that writes it.

\subsection{Small--large model collaboration in coding agents}
\label{app:related:collab}
Allocation between a cheap and an expensive model is decided after the fact by cascades, which escalate
on a verification or consistency signal \citep{frugalgpt2024,automix2024,yue2024cascade} and whose
optimality is unified with routing by \citet{dekoninck2025cascaderouting}, or in advance by learned
routers, including on-device versus cloud \citep{routellm2025,hybridllm2024} and RL-trained multi-round
variants \citep{routerr1_2025}. The decision can sit inside the acting model, calibrated for coverage
\citep{knowno2023} or post-trained jointly with solving \citep{fang2025bridging}, or be made
architectural, with the cloud writing a sketch that the device holding the private context specialises
\citep{cogenesis2024,niu2025collabsurvey}. In coding agents the same question is posed temporally
\citep{son2026swerouter} and empirically \citep{gandhi2025strongweak}. Throughout, the learned object is
a destination or a trigger. The exception composes an outbound handoff and verifies it, but for the
truthfulness of its reproduction claims rather than by re-running the task, and its frontier model
writes the patch rather than advising \citep{bhola2026scrouting}. \protege fixes the configuration we
inherit \citep{sweprotege2026}.

\subsection{Guards and sanitizers at the trust boundary}
\label{app:related:privacy}
Host-side redaction detects and substitutes sensitive spans, with reverse mapping \citep{chen2023has},
importance weighting \citep{shen2024prosan}, or inference-time search over a privacy-ordered space
\citep{zhou2025datamin}, and is scored by recall of masked spans \citep{pilan2022tab}. Our
\emph{Host sanitizer} belongs to this family, but we decline its metric, because scoring a sanitizer's output
with the detector that produced it drives the measured leak to near zero by construction. Learned guards
classify what the policy already wrote \citep{inan2023llamaguard,rebedea2023nemo}; stronger guarantees
restrict what the writer may see or send \citep{hong2024dpopt,bagdasarian2024airgap}, and contextual
integrity supplies the predicate that licenses a role-and-type target, appropriate flow rather than
absence of personal data \citep{ghalebikesabi2025ci}. Agent benchmarks establish that stated norms do
not predict behaviour under action \citep{shao2024privacylens,mireshghallah2024confaide,agentdam2025}
and that the leak lives in intermediate calls \citep{toolprivacybench2026}; our provenance inventory
descends from canaries and verbatim extraction
\citep{secretsharer2019,carlini2021extracting,carlini2023quantifying}, and bounds literal disclosure
only, not inference \citep{staab2024beyond}. Prompt injection is the adjacent threat model we disclaim
\citep{agentdojo2024,injecagent2024,debenedetti2025camel}, though SecAlign is a structural precedent for
moving a security property into the weights \citep{secalign2025}. Methods that change what the model
writes are the axis we position against
\citep{papillon2025,huang2026needtoknow,hui2026privacyr1,kim2025seal,staab2025anonymizers,dou2024selfdisclosure,zeng2026seekhelp};
one head-to-head study reports the best tradeoff for pseudonymization, judged conversationally rather
than by re-execution \citep{xiao2026saysomethingelse}.

\subsection{Tool use in LLM agents}
\label{app:related:tools}
Our action channel follows the interleaved reason-and-act loop \citep{yao2023react}, interface-dominated
agent design \citep{sweagent2024}, and the single executable channel \citep{wang2024codeact}. Tool calls
are learned by imitation on teacher corpora, graded against a reference
\citep{qin2024toolllm,patil2024gorilla,li2023apibank}; by RL on executed calls, spent on the decision
\citep{qiao2024trice,qian2025toolrl} or on what a query retrieves \citep{jin2025searchr1}; or by
filter-then-fine-tune, which divides on the filter signal. Toolformer keeps a call when its returned
result lowers a distance-weighted loss by a margin, measured against the better of not calling and
calling without the result, so a call whose mere presence helps is rejected \citep{schick2023toolformer};
TALM and STaR keep a trace only on task outcome \citep{parisi2022talm,zelikman2022star}, a rule carried
to interpreter-filtered program synthesis \citep{haluptzok2023selftaught} and repository trajectories
\citep{pan2025swegym}. Applying an outcome filter to the content of the call is nearest to us
\citep{nogueira2017task,ma2023query,andukuri2024stargate,ambati2025socratic}. Learning \emph{when} to
call is a distinct axis we do not train
\citep{asai2024selfrag,qian2025smart,zhang2025modeling,wang2025learningtoask}.

\subsection{Selection-based post-training}
\label{app:related:posttrain}
Selection keeps the correct samples \citep{yuan2023rft} or a reward-ranked fraction \citep{dong2023raft}
and can iterate \citep{zelikman2022star,gulcehre2023rest}, though the loop saturates and then overfits
\citep{restem2023}, which is why we take a single pass. DPO's reference-anchored objective \citep{dpo2023} constrains
the gap within a pair and not the level. The preferred likelihood can fall while the margin widens
\citep{pal2024dpop,rafailov2024rtoq,xiao2024caldpo}, displaced mass lands outside the pair
\citep{razin2025displacement,yan2025threed}, and the effect is worst for near-duplicates, which our
pairs are. Two caveats cut the other way. Ranking accuracy does track win rate near the reference
\citep{chen2024notrankings}, and off-policy targets move mass slowly \citep{tajwar2024onpolicy}, an
objection that applies to our own teacher-authored supervision. Training a verifier on the discarded
negatives is the principled counterargument \citep{hosseini2024vstar}, and adding a likelihood term to
the pairwise loss reaches a related conclusion \citep{pang2024irpo}; the former addresses correctness
negatives, not ones differing by a handful of private tokens. Our teacher rewrite descends from
critique-and-revise \citep{bai2022cai}, contrastive construction \citep{yang2024rlcd}, and mistake
rewriting \citep{an2023lema}, none of which checks that the revision still executes. Mixing a second
objective against narrow-tuning regression is established practice
\citep{ouyang2022instructgpt,scialom2022rehearsal} and the risk it guards is documented
\citep{qi2024finetuningcompromises,betley2025emergent}. Safe RLHF keeps two criteria separate rather
than collapsing them \citep{dai2024saferlhf}, and safety tuning past a point yields \emph{exaggerated
safety} \citep{bianchi2024safetytuned}, the published name for the failure our replay gate exists to
prevent. We are not aware of prior work that admits a training candidate only when a task-success
predicate and a disclosure predicate hold simultaneously.

\section{Method Details}
\label{app:method}

\subsection{Algorithm}
\label{app:algorithm}
Algorithm~\ref{alg:uva} states the full pipeline of Section~\ref{sec:method} in one place. It collects
consultation states from solved rollouts, rewrites each question with the offline teacher, checks the
rewrite's disclosure and form, replays it from the identical state, and fine-tunes the student on the
survivors together with the repair trajectories.

\begin{algorithm*}[t]
\caption{Pair construction and training in utility-verified abstraction (\ours)}
\label{alg:uva}
\begin{algorithmic}[1]
\Require training tasks $\mathcal X$ with executable tests; student $\pi_\theta$; teacher $R$; expert $E$;
privacy predicate $A$ (Eq.~\ref{eq:accept}); validity check $V$; mixing weight $\lambda$
\State $\mathcal C \gets \emptyset$; $\mathcal D_{\mathsf{solve}} \gets \emptyset$ \Comment{\textbf{collect} (no host rewrite at any point)}
\For{task $x \in \mathcal X$}
  \State roll out $\pi_\theta$ on $x$ as $\tau$, injecting consultations on a fixed schedule; at each
         injected consultation, record $h$, $q^{-}$, the expert's reply, the repository snapshot, the seed,
         and the remaining budgets
  \If{$S(\tau)=1$}
    \State add the recorded consultation states of $\tau$ to $\mathcal C$
    \State add the action steps of $\tau$, with consultation turns excluded as targets, to $\mathcal D_{\mathsf{solve}}$
  \EndIf
\EndFor
\State keep one candidate per identical history in $\mathcal C$; cap candidates per task
\State $\mathcal D \gets \emptyset$
\For{$(h,q^{-}) \in \mathcal C$} \Comment{\textbf{rewrite, replay, verify}}
  \State $q^{+} \gets R(h,q^{-})$ \Comment{offline teacher: roles and types replace private entities}
  \If{$A(q^{+}) \wedge L_{\mathsf{P1}}(q^{+})<L_{\mathsf{P1}}(q^{-}) \wedge V(q^{+})$}
    \State restore the snapshot at $h$; send $q^{+}$; obtain $E$'s reply; resume $\pi_\theta$ to
           termination as $\tau^{+}$
    \If{$S(\tau^{+})=1$} \State admit $(h,q^{+})$ into $\mathcal D$ \Comment{all four conditions of $G$ hold} \EndIf
  \EndIf
  \State record the rejection stage of every non-admitted candidate
\EndFor
\State \textbf{train:} fine-tune $\theta$ on
$\mathcal L_{\mathsf{question}}(\theta;\mathcal D)+\lambda\,\mathcal L_{\mathsf{solve}}(\theta;\mathcal D_{\mathsf{solve}})$
(Eq.~\ref{eq:loss})
\end{algorithmic}
\end{algorithm*}

\subsection{Teacher prompt and the allowed abstraction structure}
\label{app:teacher}
The teacher is an offline rewriting model, distinct from the deployed expert $E$; because it sees the
private $q^{-}$, pair construction runs where the data may live, here on corpora whose ``private''
content is synthetic (SWE-smith), and locally for a genuinely private repository. Its instruction is
reproduced in full below (the \emph{Privacy prompt} baseline of \S\ref{sec:setup} prompts the base student with this same
instruction, verbatim).

\begin{promptblock}[label={prompt:teacher}]{Teacher rewriter (\texttt{\_REWRITER\_SYSTEM}), verbatim}
You rewrite a software engineer's help request so it can be sent to an outside expert WITHOUT leaking
any private code detail, while staying just as useful to answer.\\[3pt]
RULES:\\
-- Remove EVERY private identifier: file paths, module/package names, class names, function/method
names, variable/parameter names, and any verbatim code or literal values.\\
-- Do NOT replace them with opaque placeholders (no IDENT\_1, FUNC\_2, etc.). Instead describe each by
its ROLE and TYPE, e.g.\ ``the constructor of a shared auth-token object'', ``a token-generator
callable'', ``a boolean formatting option'', ``a context-manager's await hook''.\\
-- KEEP all engineering substance: the observed symptom, what was expected, the STRUCTURAL pattern of
the bug (e.g.\ ``two adjacent same-type constructor arguments are passed in swapped order''), what the
engineer already ruled out, and the exact decision they need judged.\\
-- The result MUST be answerable by an expert who cannot see the code: convey enough structure to
reason about, without naming anything private.\\
-- Keep it concise (2--5 sentences). Output ONLY the rewritten question, no preamble.
\end{promptblock}

One asymmetry between this instruction and the measurement of Appendix~\ref{app:metrics} is deliberate.
The instruction removes \emph{every} identifier, including public library and standard-library names
that the detector never charges (its allowlist). At rewrite time neither the teacher nor, later, the
student can reliably decide provenance, whether a name is public API or a repository symbol shadowing
one, so the target errs toward removal, and the utility cost of over-abstraction is policed by the
replay gate rather than by trusting the rewriter's judgment, since a rewrite that abstracts away a
load-bearing name, public or private, fails its replay and is rejected. The measurement side is
unaffected, because $L_{\mathsf{P1}}$ charges only provenance-verified private entities either way.

The validity check $V$ in Eq.~\ref{eq:gate} enforces the same structure mechanically; the residual
private identifiers are bounded by $A$ (Eq.~\ref{eq:accept}), not by $V$. It requires no opaque
placeholder token (a pattern check for \texttt{IDENT\_n}-style aliases), a nonempty question of bounded
length in plain prose with no code fence or shell command, whose closing sentence is a question or a
request.

\subsection{Admission rules}
\label{app:pairs}
The target question is the exact one used in the successful replay. A consultation carries no
attachment, so the question is the whole outbound message and question-level acceptance is
full-message acceptance. A pair is admitted only with a measurable
disclosure contrast, in which $q^{+}$ passes the fixed privacy predicate $A$ and discloses strictly fewer private
entities than $q^{-}$ under the sanitizer-independent measure of Appendix~\ref{app:metrics}; we do
\emph{not} additionally require $q^{-}$ to fail $A$. Identical pairs, pairs with no disclosure
reduction, malformed asks, missing snapshots, incomplete logs, and replays with modified tests are
rejected. Neither source success nor
replay success licenses a claim that an ask was necessary. The training-data grid of
Section~\ref{sec:experiments} uses the first $100$ to $400$ accepted pairs in collection order, and
the student of Table~\ref{tab:main} is trained on the first $300$.

\section{Privacy Measurement}
\label{app:metrics}

\subsection{Definitions and audit scope}
The reference $r_h$ is the raw machine-side view at prefix $h$, the file contents, tracebacks, and
command outputs the agent has observed, and it is never transmitted. Each transmitted question is
scored against the reference of the prefix at which it was sent. Let $I(r)$ be the allowlist-filtered
private-entity set of reference $r$, constructed below, let $W_5(r)$ be the set of token $5$-grams on
the lines of $r$ that contain code syntax (an equals sign, a bracket, a colon, or a keyword such as
\texttt{def} or \texttt{return}), with comments removed, and let $w_5(q)$ be the token $5$-grams over
all lines of the message $q$. For every policy the counts are taken on the message it transmits, which
for \emph{Host sanitizer} is $T(q)$. The P1 measure used throughout the paper is $L_{\mathsf{P1}}(q;r)$
of Eq.~\ref{eq:lp1}, the number of distinct private entities disclosed, and the outbound code overlap
is $\mathsf{OCO}(q;r)$ of Eq.~\ref{eq:oco}, the number of leaked $5$-grams. We report counts rather
than ratios, because a ratio over a small reference misranks a short abstract question by exposure;
these are unique-span counts, not transmitted-token counts.

\paragraph{Per-consultation mean and consultation rate.}
For a policy $\pi_\theta$ under a fixed expert and execution budget, let $U(\pi_\theta)=\mathbb
E_{\pi_\theta}[S(\tau)]$ be the solve probability, with expectations over tasks and rollout randomness,
and let $N(\tau)$ be the number of transmitted consultations in rollout $\tau$. The mean disclosure per
consultation under a measure $\ell\in\{L_{\mathsf{P1}},\mathsf{OCO}\}$ is
\begin{equation}
D_\ell(\pi_\theta)
=
\frac{
\mathbb E_{\pi_\theta}\!\left[
\sum_{j=1}^{N(\tau)} \ell(q_j;r_{h_j})
\right]
}{
\mathbb E_{\pi_\theta}[N(\tau)]
},
\label{eq:perconsult}
\end{equation}
where $q_j$ is the message transmitted at consultation history $h_j$; it is undefined when
$\mathbb E_{\pi_\theta}[N(\tau)]=0$. The expected total disclosure per task is
$\mathbb E_{\pi_\theta}[N(\tau)]\,D_\ell(\pi_\theta)$, and the consultation rate is
$\mathsf{Ask}(\pi_\theta)=\Pr_{\pi_\theta}(N(\tau)>0)$. Table~\ref{tab:main} reports $\mathrm{pass}@1$
as the estimate of $U$, the two means $D_{L_{\mathsf{P1}}}$ and $D_{\mathsf{OCO}}$ pooled over all
transmitted consultations, and $\mathsf{Ask}$ as Ask\%.

\paragraph{Construction of $I(r)$ and the detector-only count.}
Both classes are determined automatically. P0 spans are matched by fixed patterns for credentials,
URLs, absolute paths, and high-entropy strings, and P1 membership is decided by provenance, that is
by membership in $I(r)$. The inventory $I(r)$ is assembled from the machine side, never from a sanitizer's output. The raw
machine-side view at prefix $h$, the file contents, tracebacks, and command output the agent has
actually observed, is tokenized for candidate entities (identifiers, module and file names, and absolute
paths), and the versioned allowlist removes standard-library, builtin, and
protocol names together with tokens too short or too generic to identify anything. A span in the
payload is charged only if it is in $I(r)$, so a public API name the model could have produced from
its own parameters is not counted, and a private name the model paraphrased instead of writing is
correctly not counted either. The entity recognizer is the pseudonymizer distributed with SWE-agent
\citep{sweagent2024}, used here only to \emph{enumerate} candidates; the provenance test against
$I(r)$ is what makes the measure independent of any arm's transform, and it is the reason the
\emph{Host sanitizer} arm, whose transform is that same pseudonymizer, is not scored by its own
verdict, which would put its leak near zero by construction.

The gate of Eq.~\ref{eq:accept} uses a looser \emph{detector-only} count, which charges every span the
recognizer flags in the question, with no provenance test. It is typically larger than
$L_{\mathsf{P1}}$, because it also charges public names that happen to look private. Every disclosure
number we report is the provenance-verified count.

For \emph{Host sanitizer}, the reverse map stays local and is never transmitted. Both $W_5(r)$ and
$w_5(q)$ are taken over code lexemes, namely identifiers, numbers, multi-character operators, and
single punctuation marks, after comment stripping, and $W_5(r)$ uses only the source-like lines of the
observed context. The overlap scorer identifies source-like lines heuristically, so $\mathsf{OCO}$ is a
retention proxy rather than a precise measurement of proprietary source alone.

\subsection{Terminal-Bench release}
\label{app:tbinventory}
\paragraph{Choice of a frozen release.}
Terminal-Bench is a continuously evolving benchmark. Tasks are added, modified, and retired between
tagged releases, and the public leaderboard tracks the newest one. We evaluate on the frozen
\textbf{2.0} release, all $89$ of its tasks (the 2.1 task set is identical), for two reasons. First,
a frozen release keeps results reproducible, since it stays byte-identical, whereas a number reported against a moving
target cannot be re-derived once tasks change. Second, the comparison a newer release would buy is not available to us
anyway, because the leaderboard entries are frontier agents driving frontier models through their own
harnesses, while our measurements are of a $9$B local student under the harness and step budget of
Appendix~\ref{app:config}. Our Terminal-Bench numbers are therefore for comparison \emph{between the
policies in this paper}, all of which run under identical conditions, and are not leaderboard-comparable
in either direction. Every task in the release is assigned; tasks whose verifier cannot be scored are
reported as such rather than dropped. The private-entity inventory is built as on repair tasks, from
what the agent observed on the machine before each consultation (Appendix~\ref{app:metrics}).

\section{Implementation and Hyperparameters}
\label{app:config}
Table~\ref{tab:config} records the implementation choices and hyperparameters behind all results.

\begin{table}[t]
\centering
\footnotesize
\setlength{\tabcolsep}{4pt}
\renewcommand{\arraystretch}{1.1}
\begin{tabularx}{\linewidth}{@{}P{0.24\linewidth} Q@{}}
\toprule
Item & Value\\
\midrule
Student & Qwen3.5-9B \citep{qwen3.5}, sampled at temperature $0.8$ with thinking disabled\\
Experts & DeepSeek-V3.2 for every consulting policy; GPT-5.6 Sol and Claude Opus 4.8 for \ours only\\
Teacher & DeepSeek-V3.2, prompted as in Appendix~\ref{app:teacher}, temperature $0.3$, at most $600$ output tokens\\
Training tasks & SWE-smith \citep{swesmith2025}, whose source repositories are disjoint from the $12$ of SWE-bench Verified\\
Collection & A consultation is forced at step $10$ and every $12$ steps after it, at most five per rollout, within $75$ steps; the reply enters the trajectory and the forcing instruction does not\\
Gate & Eq.~\ref{eq:accept} with $\kappa_{\mathsf{P1}}=2$ on the detector-only count, the strict reduction, $V$ of Eq.~\ref{eq:gate}, and the replay from the snapshot at $h$\\
Training (\ours) & Supervised fine-tuning on $q^{+}$ with two repair-step examples per question, which realizes $\lambda$ of Eq.~\ref{eq:loss} as an example ratio; $339$ optimizer updates at batch $8$, three epochs; learning rate $7\times10^{-5}$; sequence length $3{,}072$ tokens; prompt cap $9{,}000$ characters\\
DPO (ablation) & The same pairs as preferences against a frozen reference, $\beta=0.1$, learning rate $5\times10^{-6}$, the same $339$ updates\\
Adaptation & LoRA of rank $16$, $\alpha=32$, dropout $0.05$, on all linear projections of the $4$-bit base, merged for serving; seed $42$\\
Hardware & Training on $8$ H100 GPUs; student inference on $16$ L40S GPUs\\
Training-data grid & Nested budgets of $100/200/300/400$ accepted pairs, $300$ for the student of Table~\ref{tab:main}; mixture ratio and epochs fixed\\
Evaluation & At most three consultations per task; step limits of $200$ on Verified and $100$ on Terminal-Bench; SWE-bench Verified, all $500$ instances, three runs with a fixed seed each; Terminal-Bench, all $89$ tasks, one run (Appendix~\ref{app:stats})\\
\bottomrule
\end{tabularx}
\caption{\textbf{Implementation choices and hyperparameters.}}
\label{tab:config}
\end{table}

\paragraph{Harness note.}
The agent's shell activates the per-instance environment used by the grader. A non-interactive shell
that never activates the test environment would silently develop and self-test against the wrong
interpreter, which would invalidate the solve numbers; ensuring this activation is a prerequisite for
reporting them.

\paragraph{Consultation affordance on both benchmarks.}
The agent has a single action channel, one fenced command block per turn, and the consultation is
offered inside it rather than as a separate tool schema. The system prompt states that a block may
carry \emph{either} a shell command \emph{or} \texttt{ask\_expert "\ldots"}, and the harness intercepts
the latter, sends the question, and returns the expert's reply as that command's output. Asking is
therefore the same kind of act as running a command, costs one step, and is timed by the policy itself;
nothing in the prompt instructs the model to ask, or when. The budget (at most three consultations per
task) is stated in the prompt and enforced by the harness.

\begin{promptblock}[label={prompt:ask}]{Consultation affordance, as it appears in every consulting policy's system prompt}
TWO KINDS OF COMMAND go in that block:\\[2pt]
1. Any shell command, run on this machine:\\[2pt]
\hspace*{3mm}\texttt{grep -rn "def resolve" /testbed/src}\\[3pt]
2. \texttt{ask\_expert}, which sends one self-contained question to the cloud engineer instead of
running anything locally. Their reply comes back as the output of that command, so you can read it and
act on it in your next step. You have three questions for this task:\\[2pt]
\hspace*{3mm}\texttt{ask\_expert "The failing test raises KeyError inside resolve(), but the only}\\
\hspace*{3mm}\texttt{assignment to that key happens in the branch the traceback says was taken.}\\
\hspace*{3mm}\texttt{What could make that key missing at that point?"}\\[3pt]
They cannot see your machine, so the question has to carry what you observed, what you expected
instead, and what you are uncertain about.
\end{promptblock}

\Cref{prompt:ask} gives the wording for repair tasks. The two benchmarks receive \emph{the same}
affordance in the same position, directly after the response-format rules, as the second of ``two
kinds of command,'' with one worked example whose surface form matches the benchmark's own work (a failing test on repair tasks, a build or
service error on terminal tasks) and with the same closing sentence, that the expert cannot see the
machine and the question must therefore carry the observed behavior, the expected behavior, and the
open uncertainty. The rest of the two system prompts differs only where the benchmarks already differ,
in the working directory, the grading contract, and the submit token.

This symmetry is a measurement requirement, not a stylistic one, because the ask rate (Ask\% in
Table~\ref{tab:main}) is one of the reported quantities, and a difference in where the affordance sits
in the prompt would be indistinguishable from a difference between the task domains. We verified it by
inspection of the rendered prompts, and we report the ask rate as measured under this shared
affordance; the consultation policies differ from each other only in what is done to the question after
the model writes it (nothing, host pseudonymization, or our trained abstraction), never in how the
option is presented.

\paragraph{Loss masks.}
The question loss covers the tokens of $q^{+}$ only, and the \texttt{ask\_expert} command that opens
the consultation is masked, so no target supervises the decision to consult. The solve loss covers the
repair-action tokens; padding, role delimiters, and tool scaffolding enter neither. The DPO baseline additionally uses
frozen-reference likelihoods and a rejected-question pass.

\paragraph{Baseline fairness.}
All learners consume the same accepted questions and the same number of optimizer updates. DPO also
trains on $q^{-}$ and scores a frozen reference, so it spends more compute per update than
Eq.~\ref{eq:loss}, which uses $q^{-}$ only to select pairs. The no-replay ablation trains on every
rewrite of the same candidate pool that passed the checks before replay, the question-only ablation
sets $\lambda=0$ at the same update budget, and Host-SFT replaces $q^{+}$ with the sanitizer's output
$T(q^{-})$ under the same recipe and gate, except for the placeholder clause of $V$, which that output
necessarily violates.

\section{Additional Results}
\label{app:results}
This appendix extends the results of Section~\ref{sec:passk}. It compares the disclosure of \ours with
that of the host sanitizer (Appendix~\ref{app:hostcompare}), gives the solve rate of the trained
student without an expert and the solve rate split by whether a rollout consulted
(Appendix~\ref{app:decomp}), locates the consultations of each policy along the agent steps
(Appendix~\ref{app:timing}), and gives the measured length and the pricing of a consultation behind
the cost analysis (Appendix~\ref{app:distributions}). Appendix~\ref{app:failure} analyzes why the
untrained consulting baselines fail and follows one of their trajectories.

\subsection{Disclosure against the host sanitizer}
\label{app:hostcompare}
A stricter reading of Table~\ref{tab:main} asks whether \ours matches the \emph{Host sanitizer}'s
disclosure without host help, at a solve rate no worse than \emph{Natural}'s. With the same expert,
\ours meets this for verbatim code, with $\mathsf{OCO}$ at $5.9$ against $6.0$ on Verified and $1.4$
against $3.6$ on Terminal-Bench, and falls short for private identifiers, with $L_{\mathsf{P1}}$ at
$1.20$ against $0.97$ and $2.35$ against $0.71$. The gate of \S\ref{sec:method} admits training
questions with at most two detector-flagged P1 spans, which bounds the detector count and leaves the
provenance count free. A deployment can therefore keep the host transform as a fail-closed floor
behind the trained policy. The two are complementary, since the policy removes the surrounding code,
which renaming cannot, and the transform replaces the residual names it recognizes.

\subsection{Solve rate by consultation and expected disclosure per task}
\label{app:decomp}
\begin{table}[t]
\centering
\footnotesize
\setlength{\tabcolsep}{6pt}
\renewcommand{\arraystretch}{1.1}
\begin{tabular}{@{}l c c@{}}
\toprule
Expert & Verified & Terminal-Bench\\
\midrule
None$^{\ast}$ (\emph{Solo}) & 25.3 & 20.2\\
\rowcolor{rowhl}None & 36.2 & 24.7\\
\rowcolor{rowhl}DeepSeek & 37.6 & 29.2\\
\rowcolor{rowhl}GPT-5.6 & 38.3 & 36.0\\
\rowcolor{rowhl}Opus 4.8 & 40.6 & 43.8\\
\bottomrule
\end{tabular}
\caption{\textbf{Solve rate of the trained student with consultation disabled}, against the untrained
student working alone and the trained student with each expert. An asterisk marks the untrained
student; every other row is the student trained by \ours, the $300$-pair student of
Table~\ref{tab:main}. Both columns are $\mathrm{pass}@1$, over the $500$ Verified instances with $3$ samples each and a
single attempt on the $89$ Terminal-Bench tasks.}
\label{tab:noexpert}
\end{table}
Table~\ref{tab:noexpert} runs the trained student of Table~\ref{tab:main} with consultation disabled,
so that every rollout works alone, and sets it against the untrained student working alone and
against the trained student with each expert. The gap between the first two rows is what the training
changes in the student's own repair, and the gap between the second row and the expert rows is what
consultation adds on top of it. Training alone raises $\mathrm{pass}@1$ from $25.3$ to $36.2$ on
Verified and from $20.2$ to $24.7$ on Terminal-Bench. Consultation then adds $1.4$ and $4.5$ points
with DeepSeek-V3.2, $2.1$ and $11.3$ with GPT-5.6 Sol, and $4.4$ and $19.1$ with Claude Opus 4.8. On
Verified, most of the gain over \emph{Solo} is therefore the training, while on Terminal-Bench the
two stronger experts add more than the training does.

Table~\ref{tab:decomp} splits $\mathrm{pass}@1$ on Verified by whether a rollout transmitted at least
one consultation, and gives the expected number of private entities a policy discloses per task, its
consultations per rollout times its $L_{\mathsf{P1}}$ (Section~\ref{sec:setting}). The split is
descriptive. A rollout consults on its own initiative, so the consulting rollouts are selected by the
state the student reached, and under both untrained policies they solve less often than the rest.
\emph{Host sanitizer} is split over the one run its disclosure columns come from
(Appendix~\ref{app:stats}).

The expert can change the outcome only of the rollouts that consult, so consulting rollouts contribute
at most Ask\% points to $\mathrm{pass}@1$, and the measured split places the gain. With DeepSeek-V3.2
as the expert of both, the non-consulting rollouts of \ours solve $512$ of $1{,}389$ against $356$ of
$1{,}428$ for \emph{Natural}, which accounts for $10.4$ of the $13.0$-point gain, and its consulting
rollouts solve $52$ of $111$ against $13$ of $72$, the remaining $2.6$ points. The consulting rollouts
of the trained student also solve more often than its non-consulting ones, $46.8$ against $36.9$,
whereas under both untrained policies they solve less often than the rest. The split is not measured
for Claude Opus 4.8, so at least $15.3-7.6=7.7$ of its $15.3$-point gain over \emph{Solo} arises in
rollouts that never consulted.

\begin{table}[t]
\centering
\footnotesize
\setlength{\tabcolsep}{3pt}
\renewcommand{\arraystretch}{1.1}
\newcommand{\twoline}[2]{\begin{tabular}[b]{@{}c@{}}#1\\#2\end{tabular}}
\begin{tabularx}{\narrowtablewidth}{@{}Q c c >{\columncolor{rowhl}}c@{}}
\toprule
 & Natural & \twoline{Host}{sanitizer} & \textbf{Ours}\\
\midrule
Ask\% & $4.8$ & $4.6$ & $7.4$\\
Calls per asking rollout & $1.40$ & $1.35$ & $1.39$\\
Consulting rollouts & $13/72$ & $2/23$ & $52/111$\\
Non-consulting rollouts & $356/1{,}428$ & $123/477$ & $512/1{,}389$\\
$\mathrm{pass}@1$, consulting & $18.1$ & $8.7$ & $46.8$\\
$\mathrm{pass}@1$, non-consulting & $24.9$ & $25.8$ & $36.9$\\
$L_{\mathsf{P1}}$ per task & $0.30$ & $0.06$ & $0.12$\\
\bottomrule
\end{tabularx}
\caption{\textbf{Solve rate by whether a rollout consulted, and expected disclosure per task}, on
SWE-bench Verified. \emph{Natural} is over its $1{,}500$ rollouts and \emph{Host sanitizer} over the
one run of $500$ behind its disclosure columns in Table~\ref{tab:main}. \emph{Ours} consults
DeepSeek-V3.2. The rollout rows count solved rollouts over all rollouts in the group, and disclosure
per task is Ask\% times the calls per asking rollout times $L_{\mathsf{P1}}$ per question.}
\label{tab:decomp}
\end{table}

\subsection{The timing of consultations}
\label{app:timing}
\begin{figure*}[t]
\centering
\includegraphics{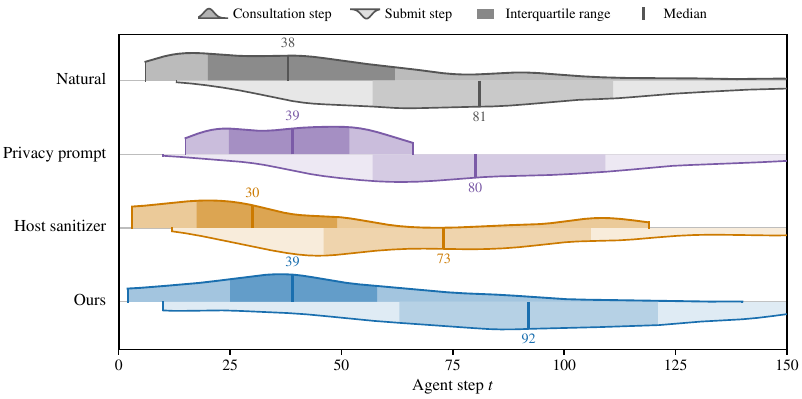}
\caption{\textbf{Consultation and submission timing per policy.} The upper half of each violin is
the distribution of the steps at which the policy consulted, and the mirrored lower half that of the step at which each submitting rollout handed in its
patch. Darker shading marks the interquartile range, and each median line carries its step. All halves
share one kernel width and peak height and span only the steps their data cover, so the violins
compare the shapes of the distributions, and Ask\% in Table~\ref{tab:main} gives how often each policy consults.}
\label{fig:timing}
\end{figure*}

\ours has no training target for when to consult (\S\ref{sec:method}), yet its Ask\% rises from $4.8$ to
$7.4$ (Table~\ref{tab:main}). Ask\% counts how many rollouts consult but not at which step, so
Figure~\ref{fig:timing} plots the steps at which each policy consults and the step at which each
rollout submits its patch, over the rollouts behind Table~\ref{tab:main}; the \emph{Ours} row uses
DeepSeek-V3.2 as the expert. We analyze only the first $150$ steps, which contain every measured
consultation and over $91\%$ of submissions. Training changes how often the student consults but not
the typical step at which it consults. The trained student consults at a median step of $39$, against
$38$ for \emph{Natural}, and submits later, at a median step of $92$ against $81$.

\subsection{Consultation length and pricing}
\label{app:distributions}
The cloud cost of Figure~\ref{fig:analysis}\figurepanel{c} multiplies, for each expert, its Ask\% from
Table~\ref{tab:main} by the consultations per asking rollout, by the tokens of a consultation, and by
the provider's published price per token. The consultations per asking rollout are $1.40$, measured on
\emph{Natural} as $101$ answered consultations over $72$ consulting rollouts, and applied to every
expert. The length is measured on the same $101$ consultations, none of them truncated. The fixed
expert system prompt is $368$ characters, and the question and the reply average $487$ and $542$
characters. We convert characters to tokens at four characters per token, since the experts tokenize
differently, which moves every point by the same factor. The prices are OpenRouter's rates per million
input and output tokens, retrieved on 2026-09-23, \$$0.269$ and \$$0.40$ for DeepSeek-V3.2, \$$2$ and
\$$10$ for GPT-5.6 Sol, and \$$5$ and \$$25$ for Claude Opus 4.8. The student's own inference is priced
at the same provider's \$$0.10$ and \$$0.15$ for Qwen3.5-9B.

\section{Baseline Failure Analysis}
\label{app:failure}
This appendix examines why the two untrained consulting baselines fall short, over the SWE-bench
Verified rollouts behind Table~\ref{tab:main}. \emph{Natural} consults in $72$ of its $1{,}500$
rollouts, with $101$ consultations, and \emph{Privacy prompt} in $7$ of its $500$, with $14$.
Disclosure is scored with the provenance inventory of Appendix~\ref{app:metrics}, resolution with the
strict rule of Table~\ref{tab:main}, and a file counts as a gold file when the reference patch edits
it. Table~\ref{tab:basecase} closes the appendix with a trajectory from \emph{Privacy prompt}.

\paragraph{Natural consults and still fails.}
$59$ of the $72$ consulting rollouts fail, and consulting rollouts solve less often than the rest,
$18.1\%$ against $24.9\%$ (Table~\ref{tab:decomp}). The failures end close to the fix. $40$ of the
$59$ submit a patch that edits a gold file and still fails the tests, $13$ edit only other files, and
$6$ submit no patch, so in most failures the student has already found where to change the code and
lacks the right change. The questions it sends leave out the evidence it holds. Before $72$ of the
$85$ failing consultations the student has seen a traceback or a failing test, yet none of the $85$
questions carries a traceback and $24$ name an error type. The questions typically restate the issue
together with the student's own hypothesis, so the expert answers from the description alone, and its
advice names a file in $11$ of the $85$. Two further patterns recur. $14$ of the $101$ consultations
ask the expert to decide whether an existing test or the issue is right, a judgement that needs the
test the expert never sees, and $7$ reproduce the worked example of \cref{prompt:ask} verbatim, a
question about a \texttt{KeyError} in \texttt{resolve()} that belongs to none of the tasks. $91$ of
the $101$ consultations disclose at least one private entity.

\paragraph{The privacy prompt still discloses.}
All $14$ consultations of \emph{Privacy prompt} disclose at least one private entity, at a mean of
$4.64$ against the $4.42$ of \emph{Natural}, and $12$ of them carry verbatim source $5$-grams. On
reading all $14$, none replaces a name with a description of its role and type, which is the core of
the instruction. The instruction changes where the disclosed names come from rather than how many
there are. Per consultation, \emph{Privacy prompt} discloses $1.2$ entities that appear only in what
the student read on the machine, against $2.4$ for \emph{Natural}, and $3.4$ that the issue text also
uses, against $2.0$. The student drops some names it found in files and restates the issue in the
issue's own names. Identifiers make up $54\%$ of its disclosed entities, and file and module names
$42\%$. Its lower disclosure per task comes from asking in $7$ of $500$ rollouts
(Section~\ref{sec:passk}), and its $6$ failing rollouts end as those of \emph{Natural} do, $4$ of them
with a patch on a gold file.

\begin{table*}[t]
\centering
\footnotesize
\renewcommand{\arraystretch}{1.15}
\begin{tabularx}{\linewidth}{@{}P{0.1\linewidth} Q@{}}
\toprule
Task & \texttt{sympy\_\_sympy-20438}, ``\texttt{is\_subset} gives wrong results''. One consultation, at
step $21$ of $116$, after the student has reproduced the failure.\\
\addlinespace[2pt]
Question & The failing test shows that \textbf{ProductSet} and \textbf{FiniteSet} with the same
elements are not considered equal (b == c is False but set(b) == set(c) is True). When
\textbf{is\_subset} is called, it falls back to comparing self.intersect(other) == self, which fails for
\textbf{ProductSet} because the intersection returns a \textbf{FiniteSet}. Additionally, the
\textbf{simplify}() of Eq(b, c) fails with AttributeError: 'Complement' object has no attribute
'equals'. What is the structural cause of this issue and how should \textbf{is\_subset} be fixed?\\
\addlinespace[2pt]
Disclosure & $L_{\mathsf{P1}}=4$, all four entities also named by the issue text; $\mathsf{OCO}=19$.\\
\addlinespace[2pt]
Expert reply & The structural cause is that \texttt{ProductSet} does not implement \texttt{\_\_eq\_\_}
or \texttt{equals} to compare element-wise with \texttt{FiniteSet} [\ldots] Immediate step: Override
\texttt{is\_subset} in \texttt{ProductSet} to compare element containment without relying on
\texttt{intersect}.\\
\addlinespace[2pt]
Outcome & The student adds that override to \texttt{sympy/sets/sets.py} and guards the
\texttt{equals} call in \texttt{sympy/core/relational.py}, and the tests fail. The reference patch
edits \texttt{sympy/core/relational.py} and the set-comparison handlers under
\texttt{sympy/sets/handlers/}, and leaves \texttt{sets.py} untouched.\\
\bottomrule
\end{tabularx}
\caption{\textbf{A \emph{Privacy prompt} consultation that discloses four entities and misdirects the
repair.} Private entities counted by $L_{\mathsf{P1}}$ are in bold; the expert reply is truncated.}
\label{tab:basecase}
\end{table*}

\paragraph{A trajectory.}
Table~\ref{tab:basecase} follows the one consultation of a \emph{Privacy prompt} rollout that shows
both failures. The question names four repository entities, every one of which the issue text also
uses, and quotes the failing expressions, although the system prompt asks for role-and-type
descriptions. The expert, answering from the description alone, proposes a mechanism, and the student
implements it in a file that the reference patch leaves untouched.

\section{Statistical Details}
\label{app:stats}
On SWE-bench Verified, every policy is run three times over all $500$ instances with a fixed seed per
run, so each row of Table~\ref{tab:main} rests on $1{,}500$ rollouts. $\mathrm{pass}@1$ is the resolved
fraction of those rollouts, that is the mean over instances of the per-instance solve rate, so it is a
count over $1{,}500$ rather than $500$.
$\mathrm{pass}@3$ is the fraction of the $500$ instances resolved by at least one of the three runs.
On Terminal-Bench, every policy is run once over all $89$ tasks. Ask\% on either benchmark is the
fraction of rollouts with at least one transmitted consultation. On Verified, the Ask\% and disclosure
columns of \emph{Privacy prompt} and \emph{Host sanitizer} come from one of their three runs, $500$
rollouts each, and those of the other consulting rows from all $1{,}500$.

Two exact tests are used. Paired comparisons of two policies on the same tasks use the exact McNemar
test with discordant counts $(b,c)$, as for the Terminal-Bench policies in Section~\ref{sec:passk}.
Consultation rates of two policies are compared by Fisher's exact test on the counts of consulting
rollouts, as in the collaboration-quality paragraph of Section~\ref{sec:passk}. A non-significant difference
is reported as inconclusive rather than as equivalence, and the conditional solve rates of
Appendix~\ref{app:decomp} are descriptive.

\section{Artifacts and Licenses}
\label{app:licenses}
We use every artifact for research evaluation, consistent with its intended use, and we release our
own code for research use under the MIT license. SWE-bench Verified \citep{swebench2024} is
distributed through the MIT-licensed SWE-bench code base; its dataset card states no separate
license, and each task repository is a public open-source project under its own license.
Terminal-Bench 2.0 \citep{terminalbench2025} is released under Apache-2.0, and the SWE-smith code
and dataset \citep{swesmith2025} under MIT. The student, Qwen3.5-9B \citep{qwen3.5}, is released
under Apache-2.0, and the weights of DeepSeek-V3.2 \citep{deepseekai2025deepseekv32} under MIT.
The experts and the teacher were accessed through commercial API providers under their terms of
service; GPT-5.6 Sol \citep{openai2026gpt56} and Claude Opus 4.8 \citep{anthropic2026opus48} are
proprietary models subject to their providers' usage policies. The mini-SWE-agent harness and the
SWE-agent pseudonymizer \citep{sweagent2024} are MIT-licensed. Issue texts and task instructions are
in English.

\end{document}